\documentclass[twoside,11pt]{article}
\usepackage{jmlr2e}
\usepackage{bm}
\usepackage{listings}
\usepackage{subcaption}
\usepackage{xcolor}
\usepackage{framed}

\usepackage{xcolor}

\colorlet{light-gray}{gray!20}

\firstpageno{1}

\begin{document}

\title{Learning to Write Notes in Electronic Health Records}

\author{\name Peter J. Liu \email peterjliu@google.com \\
       \addr Google Brain\\
       Mountain View, CA}

\maketitle

\begin{abstract}
Clinicians spend a significant amount of time inputting
free-form textual notes into Electronic Health Records (EHR) systems.
Much of this documentation work is seen as a burden,
reducing time spent with patients and contributing to clinician burnout.
With the aspiration of AI-assisted note-writing,
we propose a new language modeling task predicting the content
of notes conditioned on past data
from a patient's medical record, including patient demographics,
labs, medications, and past notes.
We train generative models using the public, de-identified MIMIC-III dataset
and compare generated notes with those in the dataset on multiple
measures. We find that
much of the content can be predicted, and that
many common templates found in notes can be learned.
We discuss how such models can be useful in supporting assistive note-writing
features such as error-detection and auto-complete.
\end{abstract}

\section{Introduction}
According to a study \citep{sinsky2016allocation}, physicians
spend nearly 2 hours doing administrative work for every hour of face-time
with patients. The most time-consuming aspect of the administrative work
is inputting clinical notes into the electronic health record (EHR) software
documenting information about the patient
including health history, assessment (e.g. diagnoses) and treatment plan.
Much of the documentation requirements are viewed as drudgery and is a
major contributor to career
dissatisfaction and burnout among clinicians. Furthermore, patient satisfaction
is affected by the intrusion of this work into time spent with patients.

The severity of the problem is such that documentation support industries
have arisen as work-arounds, ranging from dictation services, where clinicians
speak notes to be transcribed by a human or machine backend, to
scribes, human-assistants whose primary job is to write the notes. We take a
similar opinion as \citet{gellert2015rise} and view this as a sign that
EHR software usability should be improved.

Assistive-writing features for notes, such as auto-completion or error-checking,
benefit from language models. The stronger the model, the more effective such features
would likely be. Thus the focus of this paper is in building language models
for clinical notes.

Our main contributions are:

\begin{enumerate}
    \item introducing a new medical language modeling task based on
the MIMIC-III ('Medical Information Mart for Intensive Care') EHR dataset
\citep{johnson2016mimic},
\item demonstrating how to represent the multi-modal mix of structured and unstructured (text) data
found in EHRs as context for conditional language modeling;
\item proposing multiple quantitative metrics for evaluating such models
\item showing that recently developed language models 
can predict much of the content in notes while capturing their global templates;
\end{enumerate}

\section{Related Work}
The MIMIC-III dataset is comprised of de-identified electronic health records of
39,597 patients from
the intensive-care unit (ICU) of a large, tertiary care hospital. It is the most
comprehensive publicly- and freely-accessible dataset of its kind that includes
patient demographic data, medications ordered, laboratory measurements, and
of particular importance for this work, notes documented by care providers.
The release of the data and its predecessor MIMIC-II \citep{saeed2011multiparameter} has spurred many studies, predominantly focused on predicting clinical events including acute kidney injury \citep{mandelbaum2011outcome}, mortality \citep{johnson2017reproducibility}, or diagnoses and medications orders \citep{choi2016doctor}. Many other EHR datasets have also been used to predict clinical events
although they often are nonpublic or have no clinical notes.

There exists substantial prior work on utilizing clinical notes for many purposes.
\citet{friedman2004automated} extract structured output from notes in the form of 
Unified Medical Language System (UMLS) codes.
A common use of notes is to incorporate them as input to machine learning models
that predict future clinically relevant events from 
past EHR data (\citet{miotto2017deep}, \citet{rajkomar2018scalable}).
\citet{portet2009automatic} automatically summarize EHR ICU data into text form
using expert rules and templates.
\citet{jing2017automatic} train models to generate medical imaging reports
from x-rays, a type of image captioning task, 
although they do not use  data from the EHR.

We believe we are the first to focus 
on the task of building conditional language models for notes based on EHR data.
Outside of the medical domain,
language modeling has been extensively studied in the natural language
processing (NLP) community, including well-established benchmarks based on
large corpora of English sentences 
(\citet{marcus1993building}, \citet{chelba2013one}).

Conditional language modeling has also been extensively studied including for
machine translation \citep{wu2016google} and speech recognition \citep{chiu2017state},
using a class of techniques based on sequence-to-sequence learning \citep{sutskever2014sequence}.
There we model an output sequence (e.g. English words)
conditioned on an input sequence (e.g. French words).
However, most prior work relies on mapping a single
modality to text, e.g. text-to-text or audio-to-text. In our work, the data conditioned on
includes a diverse mix of both static and time-dependent data,
and sequential structured and unstructured (text) data.

\section{Methods}
\subsection{Conditional language models for clinical notes}
Language models specify a probability distribution, $P(w_1, w_2, ..., w_n)$,
over pieces of language (e.g. sentences or documents),
which can be seen as a sequence of tokens
(e.g. words), $w_1, w_2, ..., w_n$.
Using the chain-rule of probability we can write:

\[  P(w_1, w_2, \ldots w_n) = \prod_{i=1}^{n} P(w_i | w_1, ..., w_{i-1}) \]

which decomposes the joint distribution into a product of 
conditional distributions of the current token
given previous tokens. This also defines a generative model which allows
us to sample likely tokens one at a time to generate full sequences.

Conditional language models are similar except they are provided with additional
context data, $c$:

\[ P(w_1, w_2, \ldots, w_n | c)  =  \prod_{i=1}^{n} P(w_i | c, w_1, \ldots, w_{i-1}) \]

The context could be of any modality, image, text, audio, etc. For example,
in the case
of machine translation, this context is the source language sentence
to be translated. In image captioning, the context may be an image \citep{vinyals2015show}.

In this work, the sequence to be predicted is the text of the
current clinical note and context is the past data extracted from the Electronic
Health Record (EHR) for the same patient, $R$. We also augment the context
with the intended note type, $T$, and a hint of 10 tokens from the current note, $H$.
This note-context, $c=(R, T, H)$, is intended to simulate what
is known at the time a clinician begins writing a note.

Formally, we develop the model:
\[ P(D=w_1, w_2, \ldots, w_n | c =(R, T, H))\]
where $H=w_1, w_2, \ldots, w_{H_L}$ and $H_L=10$ which is smaller than $n$.
We restrict the number of note tokens to predict to $n=500$.
As is typical in machine learning, we can view this as a supervised problem,
learning the mapping $c \rightarrow D$.

\subsection{Extracting and representing context}
\label{features}
\begin{figure}[htb]
  \centering 
	\begin{framed}
        \includegraphics[width=1.0\textwidth]{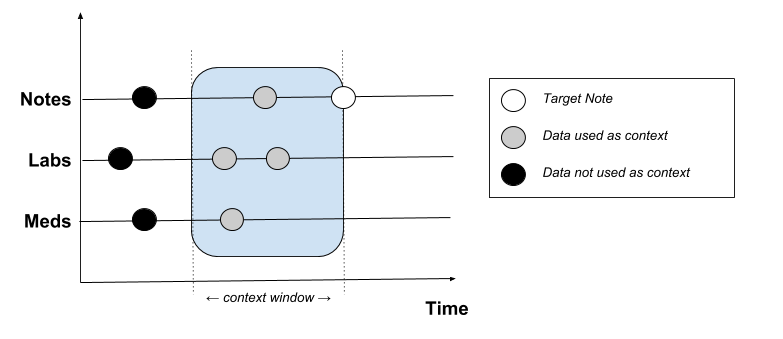}
  \end{framed}
  \caption{Schematic showing which context data is extracted from the patient record.}
  \label{fig:context} 
\end{figure} 
In our experiments we only consider context data in the 24 hours leading up
to the time of the note being written as shown in
Figure~\ref{fig:context}. We experiment with the following
context data classes:
\begin{enumerate}
	\item Demographic data ($D$): In MIMIC-III, this is static data per patient
	and found in the Patients table. We extract gender, and compute
	the age at the time of the note using the date-of-birth field.
\item Medications ($M$): from the Prescriptions table we extract names of drugs
	prescribed to the patient. These are medications prescribed within
	the context window.
\item Labs ($L$): from the LabEvents table, we extract the lab name, value,
	unit of measurement, and if available the flag saying whether it is
	abnormal. These are lab tests ordered during the context window.
\item Past notes ($N_p$): We use the text of past notes in the context window.
\end{enumerate}

The exact MIMIC-III table column names can be viewed in the
Appendix \ref{a:tables_columns}.

All the above data elements are converted to a string representation with
special delimiting tokens between data classes. Using notation
similar to Backus-Naur form, the context data is represented as:

\begin{lstlisting}[frame=single,backgroundcolor=\color{light-gray},basicstyle=\footnotesize\ttfamily]
<Context> ::= <Hint><NoteType><Demographic><MedList><LabList><NoteList>
<Demographic> ::= <Gender><Age>
<Hint> ::= first-10-tokens-of-note "<H>"
<NoteType> ::= note-type "<T>"
<Gender> ::= "M" | "F" "<G>"
<Age> ::= age-in-years "<A>"
<MedList> ::= <Medication> "<M>" | <Medication> <Delim> <MedList>
<Medication> ::= drug-name
<Delim> ::= "|"
<LabList> ::= <Lab> "<L>" | <Lab> <Delim> <LabList>
<Lab> ::= lab-name "," lab-value "," unit-of-measurement <LabFlag>
<LabFlag> ::= "abnormal" | ""
<NoteList> ::= <Note> | <Note> "<N>" <NoteList>
<Note> ::= raw-note-text
\end{lstlisting}

An example instantiation of the input is the following:

\begin{lstlisting}[frame=single,backgroundcolor=\color{light-gray},basicstyle=\footnotesize\ttfamily]
Start of note <H>Nursing/other<T>F<G>46<A>Phenylephrine|Heparin<0>
Potassium,4.1,mEq/L,|Nitrogen,4,mg/dL,abnormal<1>Progress note<N>
Another progress note
\end{lstlisting}

In our experiments we truncate the number of context tokens to 500.

\subsection{Task dataset construction}
We construct our supervised dataset in two stages.
In the first stage we randomly assign
each patient to train, validation, or test sets. In the second stage,
for every note MIMIC-III, $N$, we create a supervised learning example,
($c$, $N$).

We did not perform any patient cohort filtering as we want to use
the model for any note of any patient. However we partitioned train,
development, and test sets so that each patient would only appear in
exactly one set.
This is to prevent the model from memorizing a particular patient's history
in training and using it to advantage in predicting test notes.
The train/validation/test sizes were 1682869, 201181, 198808 respectively.

\subsection{Input and Target encoding}
\begin{figure}[htb]
  \centering 
  \begin{framed}
  \includegraphics[width=1.0\textwidth]{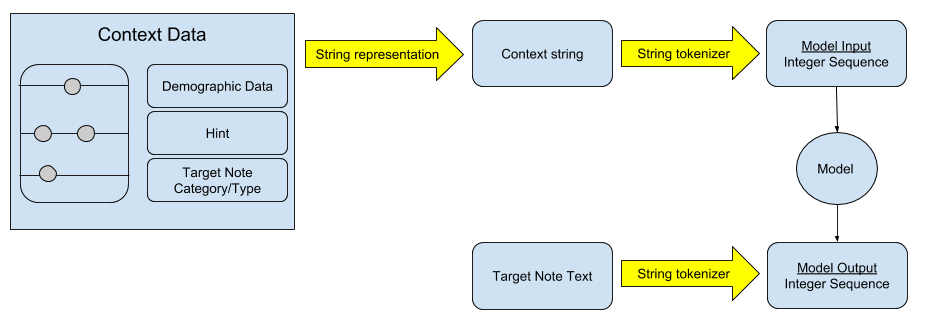}
  \end{framed}
  \caption{Schematic showing how raw data is transformed to model training data.}
  \label{fig:model-data} 
\end{figure}

We have reduced our conditional language modeling task
to that of  training a supervised model to map note-context, $c = (R, H, T)$,
a sequence of strings, 
to note-text, $N$, another sequence of strings.
Figure~\ref{fig:model-data} illustrates how the data is transformed into
input and output sequences.

Both input and target are tokenized using a sub-word tokenizer with a vocabulary of
 about 32,000, derived in
a data-driven approach as described in \citet{wu2016google}. This allows
us to transform both input and output into sequences of integers representing
sub-word tokens from the same vocabulary. In particular, we do not
pre-process note text and retain all white-space included in the original
raw notes. We found that the white-space in clinical notes is quite important
in delineating sections and removing it would reduce readability.

\subsection{Model architectures}
We begin with a very strong baseline for sequence-to-sequence learning,
called the Transformer architecture
introduced in \citet{vaswani2017attention}, achieving state-of-the-art
machine translation results.  It is an encoder-decoder architecture
without any recurrent sub-modules, in contrast to recurrent neural network 
approaches that had dominated natural language processing until its publication.

In our task the input and output sequences may be much longer than in typical
machine translation tasks where the input and output sequences are sentences.
As discussed in Section \ref{sec:results} we found that while the Transformer
encoder-decoder (T-ED) architecture performs well with shorter context, $c$,
it is unable to take advantage of greater context length.

Thus we also experiment with a recently introduced Transformer-based model
that has been shown to be effectively used for large scale (i.e. much
longer sequences)
abstractive summarization in \citet{gen-wiki}, called Transformer with
memory-compressed attention, or T-DMCA. 

Both models are implemented as part of the open-source \texttt{tensor2tensor} 
package \citep{tensor2tensor}. The hyper-parameters used are described in
Appendix \ref{a:hyper}.

\section{Evaluation}
\subsection{Evaluating the language models}

We evaluate our conditional language models on several metrics, some standard
and some specific for our task:
\begin{enumerate}
	\item \textbf{Perplexity per token (PPL)}:   A standard intrinsic metric
		used in evaluating language models is the perplexity per token
        which is a measure of how well the current token is predicted
        given previous tokens averaged over the whole sequence.
        This measure is highly local in that the predictive power beyond the next token
        is not taken into account.
		Note that perlexities from models with different vocabularies
		are not really comparable. In our case all models share the
		same vocabulary of size 32,000.
		We report log-perplexity. 
	\item \textbf{Accuracy of next token (ACC)}: The accuracy of predicting
        the next token given previous tokens. Similar to perplexity
        it is a highly local metric.
        
\item \textbf{ROUGE-1, ROUGE-2 (R1, R2)}: For a model-independent evaluation
        and a more global metric,
		we look at n-gram recall and precision statistics
		comparing the candidate generated note and the ground truth note.
		We use the commonly-used ROUGE package \citep{rouge} to compute
		ROUGE-1 (unigram) and ROUGE-2 (bigram) scores. We report
        the F1 variant which is the harmonic mean of ROUGE-P (precision)
        and ROUGE-R (recall).
        ROUGE package parameters can be found in Appendix \ref{a:rouge}.
\item \textbf{ROUGE-1 after boiler-plate removal (B-R1)}:
		We found that a significant
		amount of text in notes could be described as low-information
		content boiler-plate.
		We attempt to remove boiler-plate by removing
		text lines that are also predicted by a model trained
		to generate a note conditioned only on note-type, $T$,
		and the hint, $H$ (details provided in Appendix \ref{a:boilerplate}).
		In particular, as discussed in Section \ref{sec:templates},
		this model reliably produces the hint and many canonical
		section headings.  After boiler-plate removal, we compute the
		ROUGE metrics as usual. We report the proportion of text
		removed as boiler-plate when reporting this number.
		This allows the metric to compare sections of greater
		information content, for example, sections which are
		written as a narrative by a human, rather than auto-populated
		by an EHR template.
	\item \textbf{Sex and Age accuracy}: We use regular expressions to identify
		implied age and sex in generated notes. We then report
		the overall accuracy of the model compared to the
		ground truth found in the MIMIC-III Patients table.
		For consider age correct if it is within 1 year
		of the computed age at the time of the note.
		The regular expressions used can be found in the
		Appendix \ref{a:sex_age}.
\end{enumerate}
Note all numbers except for log-perplexity are expressed as a percentage
and bounded between 0 and 100 and higher means better. In the case of
perplexity, lower is better.

\section{Results and Discussion}
\label{sec:results}
\subsection{Template learning}
\label{sec:templates}
In our examination of model samples we noticed that commonly-used templates
in the dataset were learned and consistently employed in note generation.
This was seen even in the models conditioned on very little information.

Figure~\ref{fig:radio-note}, shows a generated note (from validation set) from a model conditioned only
on note-type ($T$) and the short hint ($H$) along with the associated ground truth.  Although
much of the content is different (e.g. age is wrong according
to the Patient table), the template was well-inferred
with appropriate section titles and formatting, and filled in with plausible values
for a radiology report.

\begin{figure} 
	\begin{subfigure}[b]{\textwidth}
		\begin{lstinputlisting}{radio_notetype_hint.txt}
		\end{lstinputlisting}
	\end{subfigure}
	\medskip
	\begin{subfigure}[b]{\textwidth}
		\begin{lstinputlisting}{radio_gt.txt}
		\end{lstinputlisting}
	\end{subfigure}
	\caption{\emph{Top:} A radiology note generated from Model 2 (from Table~\ref{tab:main-results}),
		conditioned on note-type and hint (\texttt{[**2101-7-12**] 5:44}).
		The model is able to capture the correct global structure of such notes.
	\emph{Bottom:} Ground-truth. }
  \label{fig:radio-note}
\end{figure}

We found that note-type-specific templates and styles were learned as well. Figure~\ref{fig:npn-note}
shows a nursing note whose style and structure is well emulated by the model.

Although all models were able to learn globally-coherent templates,
which can qualitatively  be assessed by looking at samples (more provided in the Appendix),
what separated the baseline and better models was
how well the content beyond the templates was predicted,
which we attempt to show through the quantitative results below.

\subsection{Quantitative Results}
\begin{table}[htpb]
  \small
  \centering
  \begin{tabular}{|l|l|l|l|l|l|l|l|l|l|l|}\hline
      &\bf{Model} & \bf{Context}  & \bf{PPL} & \bf{ACC} & \bf{R1} & \bf{R2} & \bf{B-R1} &  \bf{Sex} & \bf{Age}\\
	  \hline
	  \hline
        1 & {T-ED} & $T$  &
      1.89&	 60.4 & 19.8 & 9.1 & N/A &  58.5 &4.6\\
        2 & {T-ED} & $TH$ &
      1.79&	 62.6 &41.2 & 24.3& N/A & 63.3& 21.0\\
      3 & {T-ED} & $THD$ &
      1.77& 62.9	 &41.4 & 26.4& 31.6 &  99.9&97.8\\
      4 & {T-ED} & $THDM$ &
      1.79& 63.0&	 41.1& 26.2 & 32.3&  99.8 &97.5\\
      5 & {T-ED} & $THDML$ &
      1.81& 62.7&	 39.8& 25.1& 31.4&  99.8  &97.7\\
      6 & {T-ED} & $THDMLN_p$ &
      1.86& 62.2&	 40.5& 25.5& 32.3&  99.5 &97.4\\

      \hline

	  7 & {T-DMCA} & $THDM$ &
      1.76&	62.8 &43.1 & 27.2&  34.3&  99.7 &96.3\\
      8 & {T-DMCA} & $THDML$ &
      1.76&63.2 &	43.1 & 27.3& 34.5 & 99.8 &95.8\\
      9 & {T-DMCA} & $THDMLN_p$ &
      1.76&63.2 & 	44.6& 28.5& 36.8   & 99.9 &95.8\\
	  \hline
  \end{tabular}
  \caption{Quantitative results for model architectures and EHR contexts used in experiments.}
  \label{tab:main-results} 
\end{table}

Table~\ref{tab:main-results} shows all metrics for all models trained in
our experiments. We analyze varying two primary dimensions:
\begin{enumerate}
	\item the context data extracted from the patient's record and used as input:
		in addition to $T$, we study the effect of adding the hint ($H$),
		demographic data ($D$), medications ($M$), lab-results ($L$),
		and previous notes ($N_p$).
	\item the model architecture used: Transformer encoder-decoder (T-ED),
		or with memory-compressed attention (T-DMCA).
\end{enumerate}

\subsubsection{T-ED experiments}
In the first set of experiments we look at the effect
of including more context data on performance using the Transformer
encoder-decoder architecture. We observe that overall performance without
the hint is relatively poor, effectively being an unconditional language
model. Simply providing the first 10 sub-word tokens reveals a lot about the
note, perhaps giving a strong hint of the template to be used in the note.

As expected the accuracy of demographics (sex/age) is roughly equivalent to random
guessing without providing the demographic context ($D$) to the models.
On this metric, Model~2 has no $D$ but does better than without the
hint (Model~1) because occasionally the age/gender is revealed in the first 10 tokens.
In all models with $D$ provided, sex accuracy is virtually 100\%. 
The age accuracy is very high at about 95\% accuracy for all models.

Interestingly,
using the same Age/Sex metrics applied to the existing notes in MIMIC-III shows
that while the sex is almost always correct,
the age in \textit{true} notes was found to be significantly less accurate (88.5\%) than
our models when comparing to the computed age from the Patient table.
We discuss using the models for finding errors in Section~\ref{sec:errors}.

Overall, we observed that the T-ED Models 3-6 were not able to take full-advantage
of the additional context provided beyond note-type, hint, and demographic data ($T+H+D$).
The model's self-reported perplexity is slightly worse, suggesting optimization issues.

We confirm a similar result as \citet{gen-wiki} that the T-ED model
has difficulty with longer sequence modeling tasks. The T-DMCA architecture
employs localized attention which results in fewer unnecessary attention weights and in
practice is easier to optimize. The model also has more capacity due to the addition of mixture-of-expert layers
\citep{shazeer2017outrageously}.
This motivated the T-DMCA experiments in the next section.

\subsubsection{T-DMCA experiments}
Models 7-9 show results for the T-DMCA model in the cases of longest note-context. The results
are improved over T-ED and the best model is the one with the most context provided, showing
the architecture can appropriately take advantage of more data, in contrast to the T-ED model.

Overall the T-DMCA with most context performs the best on all metrics, except for Age accuracy,
which we attribute to notes sometimes having age contradicting the Patient table-derived age.
The perplexity and next-token accuracy metrics show a smaller relative improvement due to their local
nature; predicting the very next token often does not require taking into account long-range dependencies.
Whereas the ROUGE-based scores show larger relative improvement due to the compounding of errors
when predicting the entire note.

Comparing Models 3 and 9 on the B-R1 and R1 metrics
we see a much greater relative improvement of 16.5\% vs 7.7\%, respectively, 
suggesting the best model is more accurately predicting non-template words, i.e. meaningful
content.

Figure~\ref{fig:npn-note} shows a full sample nursing progress note
about a newborn patient generated from the best Model 9. It has inferred the proper
global structure of such notes with identical section headings
as the ground truth note. Within each section there is a high overlap
in vocabulary used and the style is well emulated. 

\begin{figure} 
	\begin{subfigure}[b]{\textwidth}
		\begin{lstinputlisting}{npn_moe.txt}
		\end{lstinputlisting}
	\end{subfigure}
	\medskip
	\begin{subfigure}[b]{\textwidth}
		\begin{lstinputlisting}{npn_gt.txt}
		\end{lstinputlisting}
	\end{subfigure}
	\caption{\emph{Top:} (Nursing) note was generated from Model 9 (from Table~\ref{tab:main-results}). The hint provided was \texttt{NPN 0700 ... Infant remains in}.  \emph{Bottom:} ground-truth note. The section headings are identical,
    though in slightly different order. Within each section there is a high overlap in words and style.}
  \label{fig:npn-note}
\end{figure}

\subsection{Detection of errors in notes}
\label{sec:errors}
As discussed in Section~\ref{sec:results}, our models conditioned
on demographic information are able to generate the appropriate age based
on structured data given as input to the model.
For example, Figure~\ref{fig:age-error} shows the same segment from
generated and actual notes where only the generated note is correct.
This could be considered as an error in the original note,
and we consider how the model could be used to detect and prevent
errors in notes.

To demonstrate error detection in existing notes, we corrupted select
notes by replacing certain drug names with random ones to simulate errors. We then
took one of our conditional language models trained to use medications as context to compute
the likelihood of the current token given previous tokens, iteratively,
for every word in the note. Words with very low likelihood
compared to surrounding context were labeled as possible errors.

\begin{figure}
	\begin{subfigure}[t]{1.0\textwidth}
		\begin{framed}
			\includegraphics[width=1\linewidth]{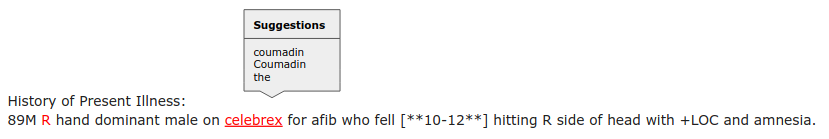}
		\end{framed}
		\caption{coumarin was replaced with celebrex,
		and the model suggests coumadin, which is an alternate name}

	\end{subfigure}
	\hfill

	\begin{subfigure}[t]{1.0\textwidth}
		\begin{framed}
			\includegraphics[width=1\linewidth]{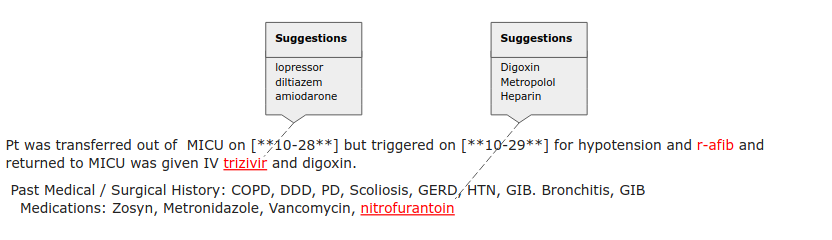}
		\end{framed}
		\caption{ amiodarone is one of correct suggestions for trizivir, while
		suggestions for nitrofurantoin are related heart medications to the original, lasix }
	\end{subfigure}
	\hfill

	\begin{subfigure}[t]{1.0\textwidth}
		\begin{framed}
			\includegraphics[width=1\linewidth]{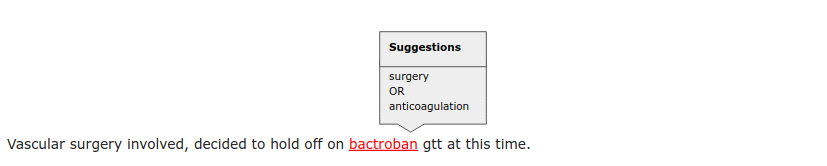}
		\end{framed}
		\caption{the original drug was heparin, an anticoagulation drug}
	\end{subfigure}
	\caption{Three corrupted snippets each from a different test note from MIMIC-III.
		Words highlighted in red are deemed unlikely under the model conditioned
		on medications. Underlined words were the result of 
		artificial corruption and the suggested model corrections are shown.
		}
	\label{fig:errors}
\end{figure}

Figure~\ref{fig:errors} shows a sample of note segments from the
test set with low likelihood words highlighted in red.
For the words we corrupted we display the top suggestions for replacement.
In most cases, only the corrupted word is highlighted and suggestions
either include the original word, or related terms.

\begin{figure} 
	\begin{subfigure}[b]{0.45\textwidth}
		\begin{lstinputlisting}{error_tdmac.txt}
		\end{lstinputlisting}
	\end{subfigure} 
    \hfill
	\begin{subfigure}[b]{0.45\textwidth}
		\begin{lstinputlisting}{error_gt.txt}
		\end{lstinputlisting}
	\end{subfigure}
    \caption{(Left) Segment of a note describing age and gender of a patient generated from Model 9 from Table~\ref{tab:main-results}. The correct age according to the Patient table is 45.
	(Right) Corresponding note segment found in MIMIC-III.}
  \label{fig:age-error}
\end{figure}

\subsection{Note auto-complete}
\label{sec:autocomplete}
Given that our models are trained to predict the next token given prior context they naturally
lend themselves to an auto-complete feature, where the next word, sentence, or more are predicted and
presented to the user as suggestions, as commonly found in software code editors \citep{raychev2014code}. 

Table~\ref{tab:top5} shows predictions from a model
for each token of a segment of a note conditioned on previous tokens and note-context. As can be seen,
age and gender are correctly predicted as well as many medical and non-medical terms. One
would expect that the more accurate such predictions are, especially when predicting long sequences of future tokens, 
the greater time saved in typing notes given a well-designed auto-complete feature.  

\begin{table}[htpb]
    \small
	\begin{tabular}{|l|l|}
        \hline
	\textbf{Token} & \textbf{Top-5 predictions (probability, token)} \\
        \hline
        History& (0.94, History), (0.00095, Past), (0.00044, Allergies), (0.00014, Date), (0.00013, HISTORY) \\
        of&      (0.92, of), (0.00044, OF), (0.00017, \textbackslash n), (7.2e-05, or), (6.6e-05, from) \\
        Present& (0.94, Present), (0.0005, present), (0.00029, Birth), (0.00016, Invasive), (0.00016, Illness) \\
        Illness& (0.94, Illness), (0.00025, Complaint), (0.00024, ILLNESS), (0.00015, Present), (0.00012, HPI) \\
        &(0.91, :\textbackslash n), (0.0072, :\textbackslash n[**), (0.0055, :\textbackslash n ), (0.0022, :\textbackslash n), (0.00052, :\textbackslash n\textbackslash n) \\
        67 &(0.27, 67), (0.15, 67), (0.071, This), (0.064, 66), (0.041, The) \\
        F  &(0.47, F), (0.3, yo), (0.057, y), (0.034, yoF), (0.0092, year) \\
        bic &(0.32, s), (0.069, with), (0.068, who), (0.057, pedestrian), (0.04, struck) \\
        ycli &(0.96, ycli), (0.028, ycl), (0.00051, restrained), (0.00034, ped), (0.00032, lo) \\
        st  &(0.98, st), (0.0023, ng), (0.001, nger), (0.001, ne), (0.00048, vs) \\
        hit &(0.4, struck), (0.39, vs), (0.017, driver), (0.016, hit), (0.011, versus) \\
        by  &(0.86, by), (0.016, in), (0.0062, a), (0.0048, head), (0.0047, at) \\
        car &(0.68, car), (0.13, a), (0.02, truck), (0.018, vehicle), (0.014, auto) \\
        , &(0.13, , ), (0.087, . ), (0.081, while), (0.063, at), (0.052, , +) \\
        unknown &(0.083, thrown), (0.054, GCS), (0.046, struck), (0.041, no), (0.03, unrestrained) \\
        LOC &(0.46, LOC), (0.077, loss), (0.048, speed), (0.028, details), (0.021, loc) \\
        , &(0.47, , ), (0.11, . ), (0.055, at), (0.054, .  ), (0.015, , +) \\
        was &(0.23, GCS), (0.042, intubated), (0.031, found), (0.026, no), (0.02, transferred) \\
        wearing &(0.13, taken), (0.055, found), (0.054, brought), (0.04, intubated), (0.029, struck) \\
        helmet &(0.38, a), (0.24, helmet), (0.065, car), (0.034, an), (0.015,  [**) \\
        but &(0.12, and), (0.053, at), (0.041, upon), (0.037, while), (0.033, but) \\
        this &(0.18, was), (0.033, found), (0.029, hit), (0.029, had), (0.022, did) \\
        was &(0.42, was), (0.056, time), (0.042, is), (0.019, AM), (0.018, morning) \\
        found &(0.1, not), (0.081, unwitnessed), (0.054, witnessed), (0.042, a), (0.021, an) \\
        separated &(0.22, to), (0.12, by), (0.11, on), (0.061, down), (0.037, at) \\
        from &(0.35, from), (0.23, by), (0.058, . ), (0.039, and), (0.035, on) \\
        patient &(0.41, her), (0.11, the), (0.085, car), (0.04, vehicle), (0.029, a) \\
        . & ( 0.14, . ), (0.12, and), (0.11, ), (0.099, .  ), (0.061, , ) \\
        GCS &(0.25, She), (0.09, Pt), (0.06, Patient), (0.03, Per), (0.028, No) \\
        of &(0.23, was), (0.19, 15), (0.074, at), (0.065, of), (0.043, on) \\
        15 &(0.35, 15), (0.14, 3), (0.11, 14), (0.042, 13), (0.04, 8) \\
        when &(0.31, at), (0.19, on), (0.053, .  ), (0.048, , ), (0.039, in) \\
        EMS &(0.33, \textbackslash n), (0.23, EMS), (0.18, she), (0.031, the), (0.013, he) \\
        arrived &(0.78, arrived), (0.085, was), (0.033, called), (0.0068, arrival), (0.0056, found) \\
        . &( 0.24, .  ), (0.15, , ), (0.081, and), (0.079, at), (0.068, . ) \\
        Taken &(0.14, She), (0.1, Pt), (0.076, Patient), (0.036, Upon), (0.035, Intubated) \\
        to &(0.85, to), (0.013, emergently), (0.0093, by), (0.0041, for), (0.0036, directly) \\
        OSH &(0.55,  [**), (0.22, OSH), (0.063, an), (0.018, the), (0.011, outside) \\
        where &(0.57, where), (0.13, , ), (0.077, and), (0.021, ED), (0.019, .  ) \\
        she &(0.46, she), (0.067, CT), (0.058, head), (0.05, a), (0.049, GCS) \\
        was &(0.67, was), (0.13, had), (0.039, received), (0.0062, became), (0.0056, underwent) \\
        found &(0.3, intubated), (0.16, found), (0.052, noted), (0.032, awake), (0.019, alert) \\
        to &(0.87, to), (0.0042, on), (0.0027, with), (0.0021, by), (0.0017, in) \\
        have &(0.65, have), (0.24, be), (0.00084, develop), (0.00068, \textbackslash n), (0.00064, to) \\
        a &( 0.27, a), (0.077, multiple), (0.034, \textbackslash n), (0.027, GCS), (0.025, SAH) \\
        splenic &(0.11, right), (0.089, large), (0.075, GCS), (0.068, small), (0.066, \textbackslash n) \\
        laceration &(0.65, laceration), (0.077, lac), (0.043, hematoma), (0.039, bleed), (0.015, contusion) \\
        requiring &(0.24, , ), (0.21, and), (0.16, .  ), (0.11, with), (0.029, . ) \\
        \hline
    \end{tabular}
    \caption{Sub-word tokens, one per line, and top-5 predictions with associated probabilities
    from one of our language models.}
    \label{tab:top5}
\end{table}


\section{Limitations and Future Work}

In our experiments we restricted our context data within 24 hours before the note
to limit input sequence length for performance reasons, although conceivably
additional past context would be informative. Due to the intensive nature
of the ICU, many events may
occur within a 24-hour period, although for non-ICU datasets a greater window
of context should be used.

Furthermore, more columns
from more tables found in MIMIC-III could be added as note-context using
the same procedure described to improve the language model, although this
was not explored.

In many cases, the maximum context provided by the EHR is insufficient 
to fully predict the note. The most obvious case is the lack of imaging
data in MIMIC-III for radiology reports. For non-imaging notes
we also lack information about the latest patient-provider interactions.
Future work could attempt to augment the note-context with data beyond the EHR,
e.g. imaging data, or transcripts of patient-doctor interactions.

Although we discussed error-correction and auto-complete features
in EHR software, their effects
on user-productivity were not measured in the clinical context, which we leave
as future work.

\section{Conclusion}
We have introduced a new language modeling task for clinical notes based on EHR data and 
showed how to represent the multi-modal data  context to the model. We proposed
evaluation metrics for the task and presented encouraging results showing
the predictive power of such models. We discussed how such models 
could be useful in sophisticated spell-checking and auto-complete features,  
potentially assisting with the burden of writing accurate
clinical documentation.


\acks{We thank Kun Zhang for assistance in data preparation; Claire Cui
for much feedback and suggesting evaluation metrics; Kai Chen and Jeff Dean
for reviewing the manuscript. We also thank the Tom Pollard, Roger Mark,
and Alistair Johnson
from the MIT Lab for Computation Physiology for approving the inclusion
of selected de-identified notes from MIMIC-III in this publication.}

\clearpage

\bibliography{main}

\begin{thebibliography}{24}
\providecommand{\natexlab}[1]{#1}
\providecommand{\url}[1]{\texttt{#1}}
\expandafter\ifx\csname urlstyle\endcsname\relax
  \providecommand{\doi}[1]{doi: #1}\else
  \providecommand{\doi}{doi: \begingroup \urlstyle{rm}\Url}\fi

\bibitem[Chelba et~al.(2013)Chelba, Mikolov, Schuster, Ge, Brants, Koehn, and
  Robinson]{chelba2013one}
Ciprian Chelba, Tomas Mikolov, Mike Schuster, Qi~Ge, Thorsten Brants, Phillipp
  Koehn, and Tony Robinson.
\newblock One billion word benchmark for measuring progress in statistical
  language modeling.
\newblock \emph{arXiv preprint arXiv:1312.3005}, 2013.

\bibitem[Chiu et~al.(2017)Chiu, Sainath, Wu, Prabhavalkar, Nguyen, Chen,
  Kannan, Weiss, Rao, Gonina, et~al.]{chiu2017state}
Chung-Cheng Chiu, Tara~N Sainath, Yonghui Wu, Rohit Prabhavalkar, Patrick
  Nguyen, Zhifeng Chen, Anjuli Kannan, Ron~J Weiss, Kanishka Rao, Katya Gonina,
  et~al.
\newblock State-of-the-art speech recognition with sequence-to-sequence models.
\newblock \emph{arXiv preprint arXiv:1712.01769}, 2017.

\bibitem[Choi et~al.(2016)Choi, Bahadori, Schuetz, Stewart, and
  Sun]{choi2016doctor}
Edward Choi, Mohammad~Taha Bahadori, Andy Schuetz, Walter~F Stewart, and Jimeng
  Sun.
\newblock Doctor ai: Predicting clinical events via recurrent neural networks.
\newblock In \emph{Machine Learning for Healthcare Conference}, pages 301--318,
  2016.

\bibitem[Friedman et~al.(2004)Friedman, Shagina, Lussier, and
  Hripcsak]{friedman2004automated}
Carol Friedman, Lyudmila Shagina, Yves Lussier, and George Hripcsak.
\newblock Automated encoding of clinical documents based on natural language
  processing.
\newblock \emph{Journal of the American Medical Informatics Association},
  11\penalty0 (5):\penalty0 392--402, 2004.

\bibitem[Gellert et~al.(2015)Gellert, Ramirez, and Webster]{gellert2015rise}
George~A Gellert, Ricardo Ramirez, and S~Luke Webster.
\newblock The rise of the medical scribe industry: implications for the
  advancement of electronic health records.
\newblock \emph{Jama}, 313\penalty0 (13):\penalty0 1315--1316, 2015.

\bibitem[Jing et~al.(2017)Jing, Xie, and Xing]{jing2017automatic}
Baoyu Jing, Pengtao Xie, and Eric Xing.
\newblock On the automatic generation of medical imaging reports.
\newblock \emph{arXiv preprint arXiv:1711.08195}, 2017.

\bibitem[Johnson et~al.(2016)Johnson, Pollard, Shen, Li-wei, Feng, Ghassemi,
  Moody, Szolovits, Celi, and Mark]{johnson2016mimic}
Alistair~EW Johnson, Tom~J Pollard, Lu~Shen, H~Lehman Li-wei, Mengling Feng,
  Mohammad Ghassemi, Benjamin Moody, Peter Szolovits, Leo~Anthony Celi, and
  Roger~G Mark.
\newblock Mimic-iii, a freely accessible critical care database.
\newblock \emph{Scientific data}, 3:\penalty0 160035, 2016.

\bibitem[Johnson et~al.(2017)Johnson, Pollard, and
  Mark]{johnson2017reproducibility}
Alistair~EW Johnson, Tom~J Pollard, and Roger~G Mark.
\newblock Reproducibility in critical care: a mortality prediction case study.
\newblock In \emph{Machine Learning for Healthcare Conference}, pages 361--376,
  2017.

\bibitem[Lin(2004)]{rouge}
Chin-Yew Lin.
\newblock Rouge: A package for automatic evaluation of summaries.
\newblock In \emph{Text summarization branches out: Proceedings of the ACL-04
  workshop}, volume~8. Barcelona, Spain, 2004.

\bibitem[Liu et~al.(2018)Liu, Saleh, Pot, Goodrich, Sepassi, Kaiser, and
  Shazeer]{gen-wiki}
Peter~J. Liu, Mohammad Saleh, Etienne Pot, Ben Goodrich, Ryan Sepassi, Lukasz
  Kaiser, and Noam Shazeer.
\newblock Generating {W}ikipedia by summarizing long sequences.
\newblock In \emph{International Conference on Learning Representations}, 2018.

\bibitem[Mandelbaum et~al.(2011)Mandelbaum, Scott, Lee, Mark, Malhotra, Waikar,
  Howell, and Talmor]{mandelbaum2011outcome}
Tal Mandelbaum, Daniel~J Scott, Joon Lee, Roger~G Mark, Atul Malhotra,
  Sushrut~S Waikar, Michael~D Howell, and Daniel Talmor.
\newblock Outcome of critically ill patients with acute kidney injury using the
  akin criteria.
\newblock \emph{Critical care medicine}, 39\penalty0 (12):\penalty0 2659, 2011.

\bibitem[Marcus et~al.(1993)Marcus, Marcinkiewicz, and
  Santorini]{marcus1993building}
Mitchell~P Marcus, Mary~Ann Marcinkiewicz, and Beatrice Santorini.
\newblock Building a large annotated corpus of english: The penn treebank.
\newblock \emph{Computational linguistics}, 19\penalty0 (2):\penalty0 313--330,
  1993.

\bibitem[Miotto et~al.(2017)Miotto, Wang, Wang, Jiang, and
  Dudley]{miotto2017deep}
Riccardo Miotto, Fei Wang, Shuang Wang, Xiaoqian Jiang, and Joel~T Dudley.
\newblock Deep learning for healthcare: review, opportunities and challenges.
\newblock \emph{Briefings in bioinformatics}, 2017.

\bibitem[Portet et~al.(2009)Portet, Reiter, Gatt, Hunter, Sripada, Freer, and
  Sykes]{portet2009automatic}
Fran{\c{c}}ois Portet, Ehud Reiter, Albert Gatt, Jim Hunter, Somayajulu
  Sripada, Yvonne Freer, and Cindy Sykes.
\newblock Automatic generation of textual summaries from neonatal intensive
  care data.
\newblock \emph{Artificial Intelligence}, 173\penalty0 (7-8):\penalty0
  789--816, 2009.

\bibitem[Rajkomar et~al.(2018)Rajkomar, Oren, Chen, Dai, Hajaj, Hardt, Liu,
  Liu, Marcus, Sun, et~al.]{rajkomar2018scalable}
Alvin Rajkomar, Eyal Oren, Kai Chen, Andrew~M Dai, Nissan Hajaj, Michaela
  Hardt, Peter~J Liu, Xiaobing Liu, Jake Marcus, Mimi Sun, et~al.
\newblock Scalable and accurate deep learning with electronic health records.
\newblock \emph{npj Digital Medicine}, 1\penalty0 (1):\penalty0 18, 2018.

\bibitem[Raychev et~al.(2014)Raychev, Vechev, and Yahav]{raychev2014code}
Veselin Raychev, Martin Vechev, and Eran Yahav.
\newblock Code completion with statistical language models.
\newblock In \emph{Acm Sigplan Notices}, volume~49, pages 419--428. ACM, 2014.

\bibitem[Saeed et~al.(2011)Saeed, Villarroel, Reisner, Clifford, Lehman, Moody,
  Heldt, Kyaw, Moody, and Mark]{saeed2011multiparameter}
Mohammed Saeed, Mauricio Villarroel, Andrew~T Reisner, Gari Clifford, Li-Wei
  Lehman, George Moody, Thomas Heldt, Tin~H Kyaw, Benjamin Moody, and Roger~G
  Mark.
\newblock Multiparameter intelligent monitoring in intensive care ii
  (mimic-ii): a public-access intensive care unit database.
\newblock \emph{Critical care medicine}, 39\penalty0 (5):\penalty0 952, 2011.

\bibitem[Shazeer et~al.(2017)Shazeer, Mirhoseini, Maziarz, Davis, Le, Hinton,
  and Dean]{shazeer2017outrageously}
Noam Shazeer, Azalia Mirhoseini, Krzysztof Maziarz, Andy Davis, Quoc Le,
  Geoffrey Hinton, and Jeff Dean.
\newblock Outrageously large neural networks: The sparsely-gated
  mixture-of-experts layer.
\newblock \emph{arXiv preprint arXiv:1701.06538}, 2017.

\bibitem[Sinsky et~al.(2016)Sinsky, Colligan, Li, Prgomet, Reynolds, Goeders,
  Westbrook, Tutty, and Blike]{sinsky2016allocation}
Christine Sinsky, Lacey Colligan, Ling Li, Mirela Prgomet, Sam Reynolds,
  Lindsey Goeders, Johanna Westbrook, Michael Tutty, and George Blike.
\newblock Allocation of physician time in ambulatory practice: a time and
  motion study in 4 specialties.
\newblock \emph{Annals of internal medicine}, 165\penalty0 (11):\penalty0
  753--760, 2016.

\bibitem[Sutskever et~al.(2014)Sutskever, Vinyals, and
  Le]{sutskever2014sequence}
Ilya Sutskever, Oriol Vinyals, and Quoc~V Le.
\newblock Sequence to sequence learning with neural networks.
\newblock In \emph{Advances in neural information processing systems}, pages
  3104--3112, 2014.

\bibitem[Vaswani et~al.(2017)Vaswani, Shazeer, Parmar, Uszkoreit, Jones, Gomez,
  Kaiser, and Polosukhin]{vaswani2017attention}
Ashish Vaswani, Noam Shazeer, Niki Parmar, Jakob Uszkoreit, Llion Jones,
  Aidan~N Gomez, {\L}ukasz Kaiser, and Illia Polosukhin.
\newblock Attention is all you need.
\newblock In \emph{Advances in Neural Information Processing Systems}, pages
  6000--6010, 2017.

\bibitem[Vaswani et~al.(2018)Vaswani, Bengio, Brevdo, Chollet, Gomez, Gouws,
  Jones, Kaiser, Kalchbrenner, Parmar, Sepassi, Shazeer, and
  Uszkoreit]{tensor2tensor}
Ashish Vaswani, Samy Bengio, Eugene Brevdo, Francois Chollet, Aidan~N. Gomez,
  Stephan Gouws, Llion Jones, \L{}ukasz Kaiser, Nal Kalchbrenner, Niki Parmar,
  Ryan Sepassi, Noam Shazeer, and Jakob Uszkoreit.
\newblock Tensor2tensor for neural machine translation.
\newblock \emph{CoRR}, abs/1803.07416, 2018.
\newblock URL \url{http://arxiv.org/abs/1803.07416}.

\bibitem[Vinyals et~al.(2015)Vinyals, Toshev, Bengio, and
  Erhan]{vinyals2015show}
Oriol Vinyals, Alexander Toshev, Samy Bengio, and Dumitru Erhan.
\newblock Show and tell: A neural image caption generator.
\newblock In \emph{Computer Vision and Pattern Recognition (CVPR), 2015 IEEE
  Conference on}, pages 3156--3164. IEEE, 2015.

\bibitem[Wu et~al.(2016)Wu, Schuster, Chen, Le, Norouzi, Macherey, Krikun, Cao,
  Gao, Macherey, et~al.]{wu2016google}
Yonghui Wu, Mike Schuster, Zhifeng Chen, Quoc~V Le, Mohammad Norouzi, Wolfgang
  Macherey, Maxim Krikun, Yuan Cao, Qin Gao, Klaus Macherey, et~al.
\newblock Google's neural machine translation system: Bridging the gap between
  human and machine translation.
\newblock \emph{arXiv preprint arXiv:1609.08144}, 2016.

\end{thebibliography}

\clearpage

\appendix
\section{Appendix}
\subsection{Permission to publish MIMIC-III notes in this paper}
The authors received explicit permission from the MIMIC-III team regarding
publishing the notes presented in this paper. 

\subsection{MIMIC-III Tables and columns used} \label{a:tables_columns}
\begin{itemize}
	\item 	Patients: GENDER (sex), DOB (date of birth)
	\item	Prescriptions: DRUG (drug name)
	\item	NoteEvents: CATEGORY, TEXT (raw text of note)
	\item   LabEvents: ITEMID (join key to D\_LABITEMS), VALUE, VALUEUOM, FLAG
	\item	D\_LABITEMS: LABEL (lab name)
\end{itemize}

\subsection{Matching sex and age in text} \label{a:sex_age}
We use the following regular expressions to identify implied age and sex
in text:
\begin{itemize}
	\item Age:
		\begin{verbatim} (\d+)\s*(year\s*old|y.\s*o.|yo|year\s*old|year-old|-year-old|-year old) \end{verbatim}
\item Male Sex: \begin{verbatim}
  (male|man|m|M)
  Sex:\s*(M) \end{verbatim}
  \item Female Sex: \begin{verbatim}
  (woman|female|f|F)
  Sex:\s*(F)} \end{verbatim}
\end{itemize}

\subsection{ROUGE computation } \label{a:rouge}
We use version ROUGE-1.5.5 with the following command-line options: \texttt{-m -n 2} .

To compute ROUGE scores in Table~\ref{tab:main-results} we sampled 4096
examples from the test set.

\subsection{Model hyper-parameters} \label{a:hyper}
The \texttt{tensor2tensor}  package supports
named model-type and hyper-parameter settings. We used the following common
  hyper-parameters: \begin{verbatim}
  max_length=10000,max_target_seq_length=500,max_input_seq_length=500
  \end{verbatim}
T-ED was trained for 500,000 steps. T-DMCA was trained for 150,000 steps.
We also used the model-specific hyperpameters found in Table~\ref{tab:hparams}.
\begin{table}[htpb]
	\begin{tabular}{|l|l|l|}
		\hline
		Model name & tensor2tensor model & hparam \\
		\hline
		T-ED &  \texttt{transformer} & \texttt{transformer\_base}  \\
		T-DMCA &  \texttt{transformer\_moe} & \texttt{transformer\_moe\_prepend\_8k}, \texttt{moe\_num\_experts=64}\\
		\hline
	\end{tabular}
	\caption{Model-specific hyper-parameters.}
	\label{tab:hparams}
\end{table}

For decoding notes, we used a beam search of size 2.

\subsection{Removing boiler-plate from notes} \label{a:boilerplate}
To compute the B-R1 metric we attempt to remove boilerplate from
generated and ground-truth notes in a pre-processing step before computing the ROUGE scores.  
In order to identify boiler-plate we rely on Model 2 from Table\ref{tab:main-results}. We make the assumption that
this model is incapable of predicting non-boiler-plate as it does not have access to sufficient context.
Thus any text that is identically generated by Model 2 is considered boiler-plate. We use the python
library \texttt{difflib} to remove lines that are present both in the note to be preprocessed and Model 2's generated note.

\begin{figure} 
	\begin{lstinputlisting}{sample_moe1.txt}
	\end{lstinputlisting}
	\caption{A sample echocardiogram note from Model 9 generated from the validation set.}
	\label{fig:sample1}
\end{figure}

\begin{figure} 
	\begin{lstinputlisting}{sample_moe2.txt}
	\end{lstinputlisting}
	\caption{A sample discharge summary note from Model 9 generated from the validation set.}
	\label{fig:sample2}
\end{figure}

\end{document}